# Accelerometer-Based Control of an Industrial Robotic Arm

Pedro Neto, J. Norberto Pires, *Member, IEEE*, and A. Paulo Moreira, *Member, IEEE*

*Abstract*—Most of industrial robots are still programmed using the typical teaching process, through the use of the robot teach pendant. In this paper is proposed an accelerometer-based system to control an industrial robot using two low-cost and small 3-axis wireless accelerometers. These accelerometers are attached to the human arms, capturing its behavior (gestures and postures). An Artificial Neural Network (ANN) trained with a back-propagation algorithm was used to recognize arm gestures and postures, which then will be used as input in the control of the robot. The aim is that the robot starts the movement almost at the same time as the user starts to perform a gesture or posture (low response time). The results show that the system allows the control of an industrial robot in an intuitive way. However, the achieved recognition rate of gestures and postures (92%) should be improved in future, keeping the compromise with the system response time (160 milliseconds). Finally, the results of some tests performed with an industrial robot are presented and discussed.

## I. INTRODUCTION

PROGRAMMING and control an industrial robot through the use of the robot teach pendant is still a tedious and time-consuming task that requires technical expertise. Therefore, new and more intuitive ways for robot programming and control are required. The goal is to develop methodologies that help users to control and program a robot, with a high-level of abstraction from the robot specific language. Making a robotic demonstration in terms of high-level behaviors (using gestures, speech, manual/human guidance, from visual observation of human performance, etc.), the user can demonstrate to the robot what it should do [1]-[5].

In the robotics field, several research efforts have been directed towards recognizing human gestures, recurring to vision-based systems [6], [7], motion capture sensors [2], [4], or using finger gesture recognition systems based on active tracking mechanisms [8]. Accelerometer-based gesture recognition has become increasingly popular over the last decade. The low-moderate cost and relative small size of the accelerometers make it an effective tool to detect and recognize human body gestures. Several studies have been conducted on the recognition of gestures from acceleration data using Artificial Neural Networks (ANNs) [9], [10], [11]. However, the specific characteristics of an industrial environment (colors, non-controlled sources of light, infrared radiation, etc.), the safety and reliability requirements, and the high price of some equipment has hampered the deployment of such systems in industry.

Given the above, the teach pendant continues to be the common robot input device that gives access to all functionalities provided by the robot (jog the manipulator, produce and edit programs, etc.). In the last few years the robot manufacturers have made great efforts to make user-friendly teach pendants, implementing intuitive user interfaces such as icon-based programming [12], color touch screens, a 3D joystick (ABB Robotics), a 6D mouse (KUKA Robot Group) [13], or developing a wireless teach pendant (COMAU Robotics). Nevertheless, it remains difficult and tedious to operate with a robot teach pendant, especially for non-expert users.

In this paper is proposed an accelerometer-based gesture recognition system to control an industrial robot in a natural way. Two 3-axis wireless accelerometers are attached to the human arms, capturing its behavior (gestures and postures). An ANN system trained with a back-propagation algorithm was used to recognize gestures and postures. Finally, several tests are done to evaluate the proposed system. The results of the performed tests are presented and discussed.

## II. SYSTEM OVERVIEW

### A. System Description

The demonstration cell (Fig. 1) is composed of an industrial robot MOTOMAN HP6 equipped with the NX100 controller, two 3-axis wireless accelerometers to capture human hand behaviors, and a computer running the application that manages the cell.

The 3-axis accelerometers (ADXL330, Analog Devices) are physically rated to measure accelerations over a range of at least +/- 3g, with a sensitivity of 300 mV/g and sensitivity accuracy of 10%. The accelerometers communicate with the computer via Bluetooth wireless link, reporting back data at 100 Hz (Fig. 2).

Manuscript received March 15, 2009. This work was supported in part by the European Commission's Sixth Framework Program under grant no. 011838 as part of the Integrated Project SMErobot[TM], and the Portuguese Foundation for Science and Technology (FCT), grant no. SFRH/BD/39218/2007.

Pedro Neto is a PhD student in the Industrial Robotics Laboratory - Mechanical Engineering Department of the University of Coimbra, POLO-II, Pinhal de Marrocos, 3030-788, Coimbra, Portugal; (e-mail: pedro.neto@robotics.dem.uc.pt).

J. Norberto Pires is with the Industrial Robotics Laboratory - Mechanical Engineering Department of the University of Coimbra, POLO-II, Pinhal de Marrocos, 3030-788, Coimbra, Portugal; (e-mail: jnp@robotics.dem.uc.pt).

A. Paulo Moreira is with the Institute for Systems and Robotics (ISR) - Department of Electrical and Computer Engineering of the University of Porto, Rua Dr. Roberto Frias, 4200-465, Porto, Portugal; (e-mail: amoreira@fe.up.pt).

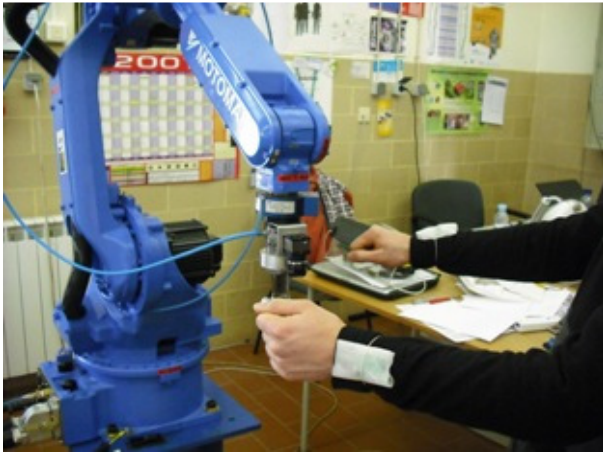

Fig. 1. Industrial robot MOTOMAN HP6 controlled by arm gestures and postures.

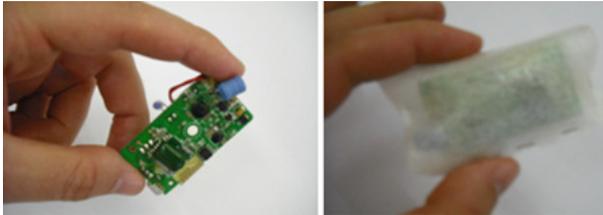

Fig. 2. 3-axis accelerometer coupled to the transmitter system. The dimensions of the device are 60x30x20 millimeters and weights 90 grams (including the battery weight). To facilitate their use, the device is placed inside a bag (right side of the image).

### B. Methodology

The 3-axis accelerometer attached to the right arm is used to recognize gestures (dynamic arm positions) and postures (static arm positions), whereas the accelerometer attached to the left arm recognizes the postures used to activate and deactivate the system (only two postures). In practice, the user should make a gesture with the right arm and at the same time use the left arm to activate or deactivate the system. When activated, the system acquires data from the accelerometer attached to the right arm, recognizes the gesture or posture and starts the robot movement. Performing a specific posture with the left arm, the robot stops. If the user never stops the robot, the robot continues the movement up to the limit of its field of operation.

An ANN system trained with a back-propagation algorithm was used to recognize gestures and postures. The ANN system has as input the motion data (extracted from the accelerometer attached to the right arm) and as output the recognized gestures and postures.

The application that manages the cell receives data from the accelerometers, interprets the received data and acts in the robot, using for this purpose the *MotomanLib*, a Data Link Library created in our laboratory to control and manage the robot remotely via Ethernet (Fig. 3). Given that the accelerometers communicate with the computer via Bluetooth, it is important to take into account the reliability of this type of communication and use it with care. The system here presented is continuously receiving data from the accelerometer attached to the left arm and if the communication fails, the robot immediately stops.

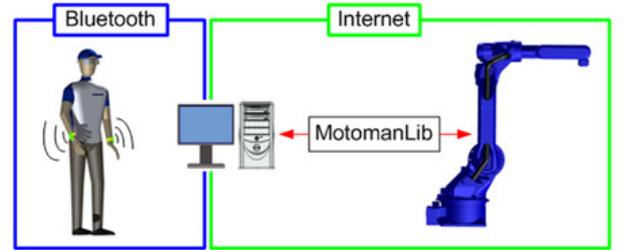

Fig. 3. A schematic representation of the platform, in terms of communication technology. The accelerometers transmit data without wires (via Bluetooth), giving a greater freedom to the user.

### III. CONTROL STRATEGY

#### A. Robot Control

The robot is controlled remotely via the Ethernet using a command that moves the robot linearly according to a specified pose increment $i = [i_1 \quad i_2 \quad i_3 \quad i_4 \quad i_5 \quad i_6]^T$. The first three components represent the robot translation along the X, Y and Z axes, respectively, whereas the last three components represent the robot rotation about the X, Y and Z axes, respectively. These components $i$ have the necessary information to control the robot. It is therefore necessary to identify them by examining the behavior of the user right arm.

In this system it is completely unnecessary to extract precise displacements or rotations, being only required to know which of the pose increment components must be activated. In a first approach, the robot control strategy was to identify translation movements and rotations of the user hand and, depending on these inputs, small pose increments were continuously sent to the robot. However, it was quickly concluded that this approach was not viable because the robot was constantly halting, presenting a high-level of vibration. The achieved solution was to send to the robot only one pose increment that will move the robot to the limit of the field of operation.

#### B. Increment calculation

According to the user right arm behavior, the robot is moved from the current pose to the limit of its field of operation, or more specifically, for a pose close to the limit of the robot field of operation. The field of operation of a 6-DOF robot manipulator is approximately a volume region bounded by two spherical surfaces. This way, it can be considered that the field of operation of the robot is bounded by two spherical surfaces (1), both with the centre coincident with the zero reference point of the robot, and where $R_{ext}$ and $R_{int}$ are respectively the radius of the external and internal spherical surface.

$$\begin{cases} x^2 + y^2 + z^2 \leq R_{ext}^2 \\ x^2 + y^2 + z^2 \geq R_{int}^2 \end{cases} \quad (1)$$

Before starting any robot movement, the "current" robot position $(x_r, y_r, z_r)$ is acquired. In order to calculate the pose increment $i$, firstly it is necessary to achieve the increment components which must be activated. This is done by referring to the acceleration values $(a_x, a_y, a_z - 1)$ that will define the robot movement direction $d$. This vector $d$ in conjugation with the "current" robot position point $(x_r, y_r, z_r)$ will be used to achieve a straight line (2) that will intersect the external spherical surface at two points (Fig. 4). In a first approach, it is considered that only the external spherical surface limits the robot field of operation.

$$(x, y, z) = (x_r, y_r, z_r) + k \cdot (d_1, d_2, d_3), k \in \Re \quad (2)$$

From (1) and (2)

$$(x_r + d_1 \cdot k)^2 + (y_r + d_2 \cdot k)^2 + (z_r + d_3 \cdot k)^2 = R_{ext}^2. \quad (3)$$

Extracting $k$ from (3), and considering only the positive value of $k$ (vector $d$ direction), the distance from the "current" robot position to the external spherical surface point (robot increment) is

$$(x, y, z) = (i_1, i_2, i_3) = k \cdot (d_1, d_2, d_3), k \in \Re^+. \quad (4)$$

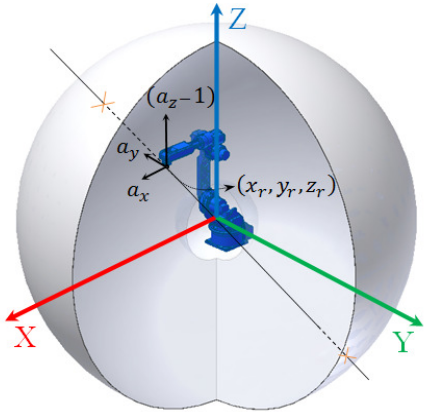

Fig. 4. The two spherical surfaces that define the robot field of operation. The "current" robot point and the acceleration vector components that will define the robot movement direction are shown in figure.

Thus, in terms of robot translation movements, the pose increment is $i = [i_1 \ i_2 \ i_3 \ 0 \ 0 \ 0]^T$. For example, if it is found that the robot should be moved along the Y axis in the negative direction, the vector $d$ becomes $(0, -1, 0)$, and then $i = [0 \ i_2 \ 0 \ 0 \ 0 \ 0]^T$. An analog approach was employed to obtain $i$ when the robot field of operation is limited by the internal spherical surface. In this case, if $k$ has no value (impossible to calculate), it means that the straight line does not intercept the internal spherical surface and it is the external spherical surface that limits the robot field of operation.

In terms of rotation increments, since it is known the robot rotation limit values and the "current" robot pose, it is easy to obtain the increments.

## IV. GESTURE AND POSTURE RECOGNITION

### A. Mode of operation

The robot moves along the X, Y and Z axes separately (robot translations). The rotation around each of the three axes is also done separately, an axis at a time.

When the accelerometer is operating in a dynamic way, the gravity components will appear mixed with the inertial components of acceleration. In order to prevent this situation, when the user makes gestures, the accelerometer must be kept horizontal (Fig. 5). Thus, it is known that the force of gravity acts along the Z axis. For example, to move the robot in the X direction, the user should move the accelerometer along the X axis, keeping it in the horizontal. Of course it is humanly impossible to keep the accelerometer exactly in the horizontal, but this is a way to have some control over the process. Thus it is relatively easier to control

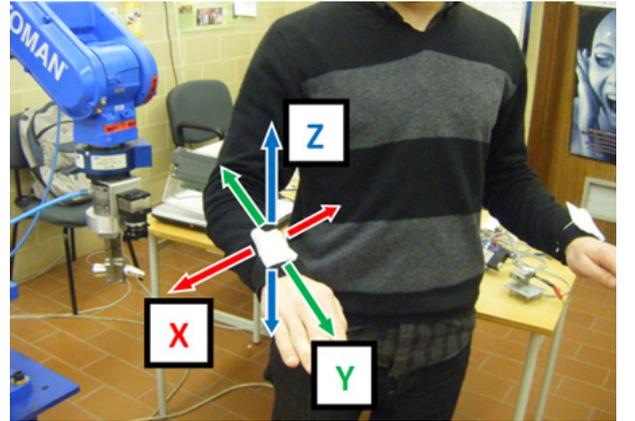

Fig. 5. The system recognizes six different gestures, the necessary to represent the robot translations (X+, X-, Y+, Y-, Z+ and Z-). In both movements the accelerometer should remain horizontally.

the acceleration due to gravity and recognize gestures.

### B. Recognition of gestures and postures

When the arm is moved in the positive X direction (X+) (Fig. 6), initially the value of acceleration $a_x$ increases because the arm begins to move and then, when the arm begins to slow the positive value of $a_x$ is converted to a negative value. This point ($a_x = 0$) marks the point of maximum speed. The acceleration $a_y$ remains near to zero and $a_z$ remains near to one because the accelerometer is held horizontally (acceleration due to gravity). A similar reasoning can be done to the other gestures (X-, Y+, Y-, Z+ and Z-).

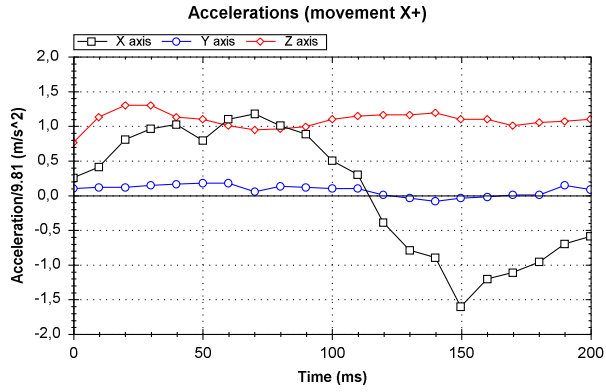

Fig. 6. Measured accelerations when the accelerometer is moved along the X axis in the positive direction (X+).

To interpret the acceleration values and recognize the right arm movements (X+, X-, Y+, Y-, Z+ and Z-), an ANN trained with a back-propagation algorithm was implemented into the system. In a first approach, the acceleration values (from the beginning of the movement to the first point of zero acceleration (maximum speed)) were used as input pattern for the ANN. However, under this approach, the robot begins to move after the user finishes the gesture, showing a significant delay from the beginning of the gesture to the moment when the robot starts to move. The aim is that the robot starts the movement almost at the same time as the user activates the robot movement (left arm) and makes a gesture with the right arm. To do this, immediately after the user activates the robot movement, the system extracts the acceleration values from the accelerometer attached to the right arm (only three measurements), identifies the gesture and sends a command to move the robot. These three measurements of acceleration will be used to recognize gestures, allowing a fast recognition. However, if the number of measured accelerations is reduced, it is more difficult to recognize a gesture and the recognition rate becomes low. Thus, these three measurements represent a compromise between the time delay and the achieved recognition rate.

In addition to the robot translations, the robot control architecture needs also to have as input six different robot rotations (Rx+, Rx-, Ry+, Ry-, Rz+ and Rz-). If the accelerometer is in free fall, it will report zero acceleration. But if the accelerometer is held horizontally, it will report an acceleration along the Z axis, the acceleration due to gravity $g$. Thus, even when the user is not accelerating the arm, a static measurement can determine the rotation of the arm (posture recognition). Analyzing figure 7-A, when the accelerometer is held horizontally, it will report an acceleration $g$ along the Z axis; $a_z \approx g$, $a_x \approx 0$, $a_y \approx 0$. When the accelerometer is rotated around the Y axis (Fig. 7-B); $a_x \approx -g$, $a_y \approx 0$, $a_z \approx 0$. On the contrary, when the accelerometer is rotated around the Y axis in the reverse direction (Fig. 7-C); $a_x \approx g$, $a_y \approx 0$, $a_z \approx 0$. A similar approach detects rotations around the X axis (Fig. 7-D and 7-E). In terms of rotation around the Z axis (Fig. 7-F, 7-G), nothing can be concluded as in both cases the gravity is along the Z axis. To solve this problem, an ANN was used to detect rotation movements around the Z axis. The ANNs were also used for recognition of postures (Rx+, Rx-, Ry+ and Ry-).

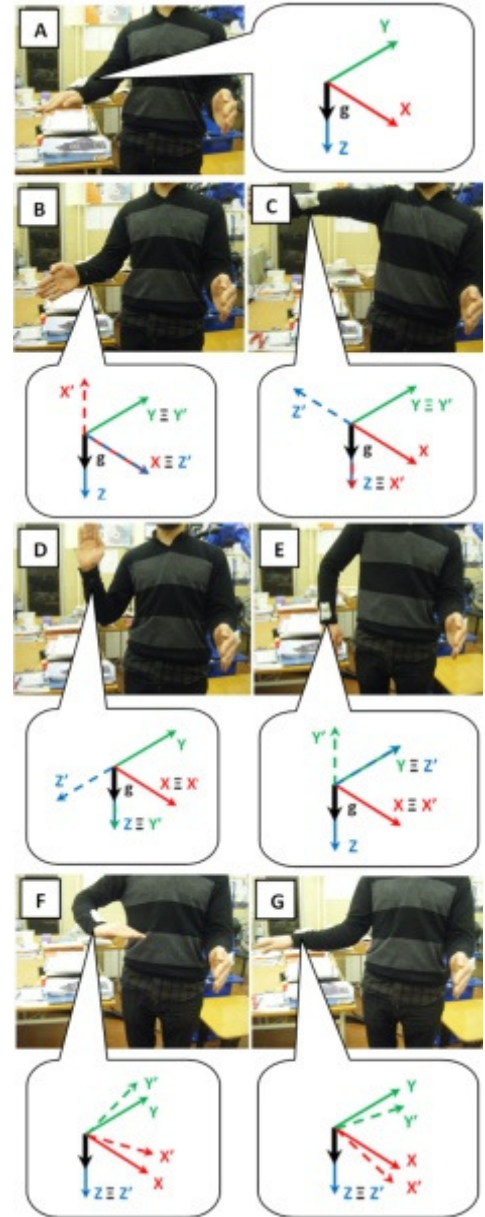

Fig. 7. Postures and gestures performed by the user right arm. A - No rotation. B - Rotation around the Y axis in the negative direction (Ry-), C - (Ry+), D - (Rx-), E - (Rx+), F - (Rz+), G - (Rz-).

In order to identify the posture of the left arm (start or stop robot movement), analyzing figure 8, when the accelerometer is held horizontally, it will report an acceleration $g$ along the Z axis in the positive direction, $a_z \approx g$, $a_x \approx 0$, $a_y \approx 0$ (the robot stops). When the accelerometer is rotated around the Y axis, $a_x \approx -g$, $a_y \approx 0$, $a_z \approx 0$ (the robot starts the movement).

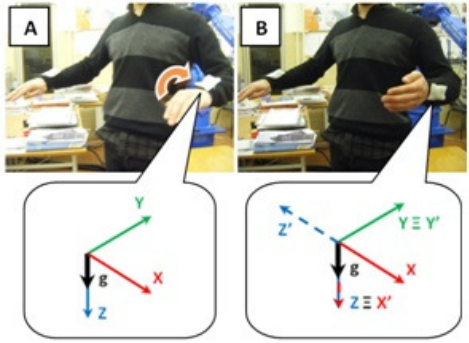

Fig. 8. Postures performed by the user left arm. A - Robot stopped. B - The robot starts the movement.

## C. Artificial Neural Network

ANNs are simple to use and present good learning capabilities. To recognize gestures and postures, an ANN trained with a backpropagation algorithm was implemented. The ANN input signals (acceleration data) are represented by a vector $x = (x_1,...,x_n)$ and the output from a neuron $i$ is given by (5), where $x_j$ is the output of neuron $j$, $w_{ij}$ is the weight of the link from neuron $j$ to neuron $i$, $\theta_i$ is the bias of neuron $i$ and $F$ is the activation function.

$$x_i = F\left(\sum_j w_{ij} \cdot x_j + \theta_i\right). \quad (5)$$

The backpropagation algorithm is used as a learning algorithm to determine the weights of the network. During the process, the weights are adjusted in order to minimize the error. The error is achieved comparing the desired output (obtained in the training phase) with the "actual" output.

A specific gesture or posture is recognized by a three-layer feed-forward ANN. The number of neurons was nine for the input layer, ten for the hidden layer and twelve for the output layer. Nine neurons in the input layer encode each gesture, three measurements of acceleration (each with three components of acceleration). Ten neurons were used in the hidden layer because after several experiments it was concluded that this solution presents a compromise between the computational time required to train the system and an acceptable recognition rate (over 90%). Finally, the twelve neurons in the output layer correspond to each different gesture and posture. Each neuron of the output layer outputs the recognition result, a numerical value between 0 and 1 (sigmoid function). If the output value is larger than or equal to 0.5, it means that the neuron detected the gesture or posture.

Finally, to define the robot increment, the recognized gestures are then transformed in the vector $d$, for example, if is detected the movement (Y+), $d = (0,1,0)$.

## D. Results

After training the system, several tests were conducted to achieve the recognition rate for each gesture and posture. The tests were conducted with two participants, and each one performed each gesture/posture 100 times. The participants were two users (P1 and P2) that trained the system before performing the tests. The results are presented in Table I.

TABLE I
RECOGNITION RATE (%)

| Gesture or Posture | User P1 | User P2 |
|---|---|---|
| X+ | 94 | 91 |
| X- | 90 | 92 |
| Y+ | 89 | 88 |
| Y- | 87 | 88 |
| Z+ | 92 | 90 |
| Z- | 89 | 92 |
| RX+ | 100 | 100 |
| RX- | 100 | 99 |
| RY+ | 100 | 100 |
| RY- | 99 | 100 |
| RZ+ | 85 | 82 |
| RZ- | 84 | 84 |
| Mean | 92 | 92 |

The recognition rate of postures is much higher than that of gestures. This discrepancy is due to the acceleration readings that provide information where the gravity components appear mixed with the inertial components of acceleration, making it harder to recognize gestures. Another problem that hinders the recognition of gestures is the necessary coordination between the arms that some users may be difficult to assimilate. Nevertheless, the experience shows that in 30 minutes any user is able to operate the system (Fig. 9).

The recognition rate depends on the samples provided during the training phase. The results presented were obtained using 30 patterns taught to the network. If the number of learning patterns is increased, the recognition rate is improved but not significantly. These 30 patterns represent a compromise between the required training time and the recognition rate. This is achieved with the ANN parameters and computer characteristics presented in Table II, where the user takes 9 minutes to train the system and 30 minutes of computational time. The recognition rate obtained by this system is in accordance with other approaches that use ANNs to recognize gestures.

TABLE II
ANN PARAMETERS AND COMPUTER CHARACTERISTICS

| Parameters | |
|---|---|
| Activation function | Sigmoid function |
| Training cycles | 100000 |
| Number of hidden neurons | 10 |
| Learning rate | 0.25 |
| Computer processor | Intel® Core™2 Duo T5600 |
| Computer RAM | 1 GB |

The system response time from the begining of the gesture to when the robot starts to move is 160 milliseconds.

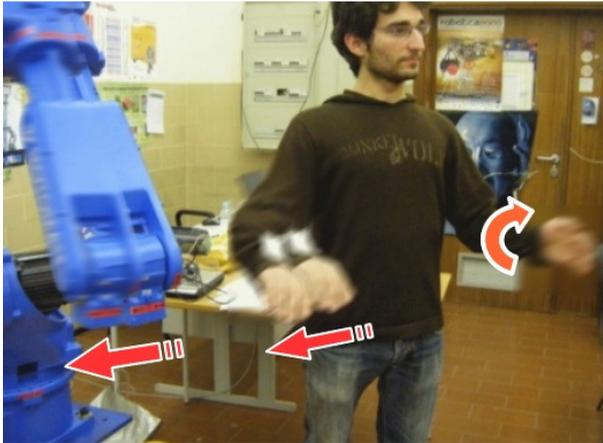

Fig. 9. The user makes a gesture and immediately, the robot starts to move in the same direction of his/her right arm.

## V. CONCLUSION AND FUTURE WORK

Due to the growing demand for natural Human Machine Interfaces and robot intuitive programming platforms, a robotic system that allows users to control an industrial robot using arm gestures and postures was proposed. Two 3-axis accelerometers were selected to be the input devices of this system, capturing the human arms behaviors. When compared with other common input devices, especially the teach pendant, this approach using accelerometers is more intuitive and easy to work, besides offering the possibility to control a robot by wireless means. Using this system, a non-expert robot programmer can control a robot quickly and in a natural way. The low price and short set-up time are other advantages of the system. Nevertheless, the reliability of the system is an important limitation to consider.

The ANN's shown to be a good choice to recognize gestures and postures, presenting an average of 92% of correctly recognized gestures and postures. The system response time (160 milliseconds) is another important factor.

Future work will build upon the improvement of the average of correctly recognized gestures. One approach might be the implementation of a gyroscope into the system, in order to separate the acceleration due to gravity from the inertial acceleration. The use of more accelerometers attached to the arms is another possibility.